\documentclass[conference]{IEEEtran}
\IEEEoverridecommandlockouts
\usepackage{cite}
\usepackage{amsmath,amssymb,amsfonts}
\usepackage{algorithmic}
\usepackage{graphicx}
\usepackage{subcaption}
\usepackage{textcomp}
\usepackage{xcolor}
\def\BibTeX{{\rm B\kern-.05em{\sc i\kern-.025em b}\kern-.08em
    T\kern-.1667em\lower.7ex\hbox{E}\kern-.125emX}}
\begin{document}

\title{Practical PCG Through Large Language Models}


\author{\IEEEauthorblockN{$^{\alpha}$Muhammad U Nasir and $^{\beta}$Julian Togelius}
\IEEEauthorblockA{$^{\alpha}$University of the Witwatersrand, South Africa, umairnasir1@students.wits.ac.za\\
$^{\beta}$New York University, USA
}
}

\IEEEoverridecommandlockouts

\IEEEpubid{\makebox[\columnwidth]{979-8-3503-2277-4/23/\$31.00~\copyright2023 IEEE \hfill} 
\hspace{\columnsep}\makebox[\columnwidth]{ }}

\maketitle

\IEEEpubidadjcol

\newcommand{\TODO}[1]{{\color{red} TODO: {#1}}}

\begin{abstract}
Large Language Models (LLMs) have proven to be useful tools in various domains outside of the field of their inception, which was natural language processing. In this study, we provide practical directions on how to use LLMs to generate 2D-game rooms for an under-development game, named \texttt{Metavoidal}. Our technique can harness the power of \texttt{GPT-3} by Human-in-the-loop fine-tuning which allows our method to create $37\%$ Playable-Novel levels from as scarce data as only 60 hand-designed rooms under a scenario of the non-trivial game, with respect to (Procedural Content Generation) PCG, that has a good amount of local and global constraints.
\end{abstract}


\section{Introduction}

There are many ways of generating game levels. While most games featuring online PCG that are actually shipped rely on domain-specific heuristic solutions, methods explored by  experimenters include evolutionary computation, constraint satisfaction, grammar expansion, and fractals~\cite{shaker2016procedural,togelius2011search}. Roguelike games in particular often feature relatively ambitious PCG methods~\cite{harris2020exploring}. Over the last decade, machine learning has turned out to be fruitfully applicable to essentially everything under the sun. This includes level generation. Researchers have explored the ways machine learning in general can be applied 
to generating levels and other types of game content~\cite{summerville2018procedural}, as well as deep learning in particular~\cite{liu2021deep}. PCG itself holds utmost importance in many important research fields, like Open-ended Learning \cite{nasir2022augmentative} or continual learning \cite{xu2018reinforced}.

It's 2023, and the new new thing that can be applied to everything under the sun is Large Language Models (LLMs), such as image generation~\cite{chang2023muse} and neural architecture search~\cite{nasir2023llmatic}. While originally developed for natural language processing, LLMs have proven effective for anything that can be expressed as sequences of tokens, including images. The versatility of LLMs go beyond what would normally be considered text completion, as they are capable of performing many tasks that would seem to require cognitive efforts from humans. Could LLMs also be useful for generating game content? Game levels, like everything else that passes through a computer, are after all just strings.

Two recent studies examine this. In one of them, GPT-2 and GPT-3 were finetuned to generate Sokoban levels. The generated levels were good and novel but, particularly for GPT-2, the dataset requirements were excessive~\cite{todd2023level}. Another study showed that special-purpose LLM architecture produced good levels for the classic platformer Super Mario Bros~\cite{sudhakaran2023mariogpt}.

In this paper we explore the possibility of using LLMs to generate levels for a game under active development, where only a limited number of levels are available, forcing us to find a data-efficient method. Furthermore, these levels are relatively large and have a nontrivial number of constraints. Our approach is to encode the constraints into the prompt and fine-tune GPT-3. To efficiently use the limited data available without overfitting we use several types of data augmentation as well as a form of bootstrapping, where novel high-quality levels are added back into the dataset.





\section{Metavoidal and Room Generation Setup}

\newcommand{\www}[1]{0.13\textwidth}
\begin{figure}[!htb]
    
    \centering
    \begin{subfigure}{\www}
        \includegraphics[width=\textwidth]{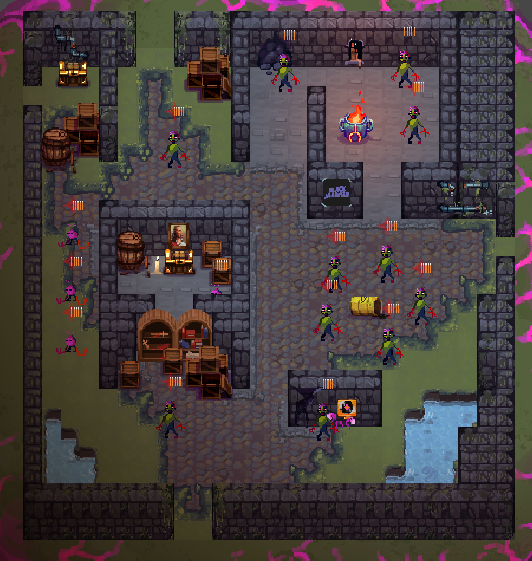}
    \end{subfigure}
    \begin{subfigure}{\www}
        \includegraphics[width=\textwidth]{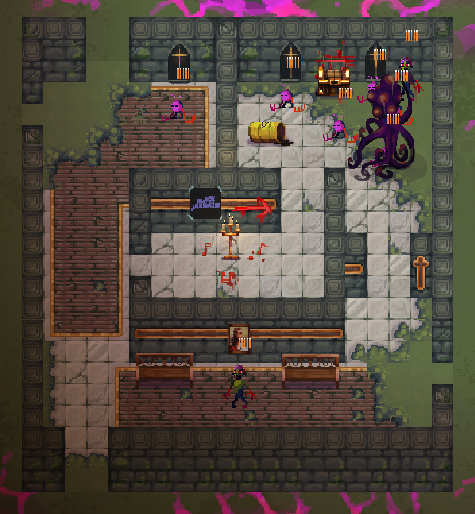}
    \end{subfigure}
    \begin{subfigure}{\www}
        \includegraphics[width=\textwidth]{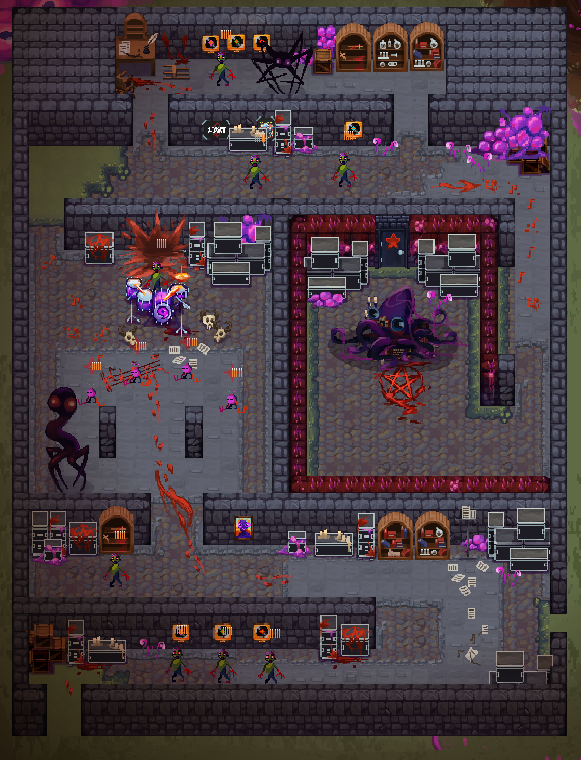}
    \end{subfigure}
    
    \caption{Levels of different sizes created by the developers with all assets on them.}
    \label{fig:OrigRooms}
\end{figure}

\begin{figure*}[h!]
    \centering
    \includegraphics[width=0.525\textwidth]{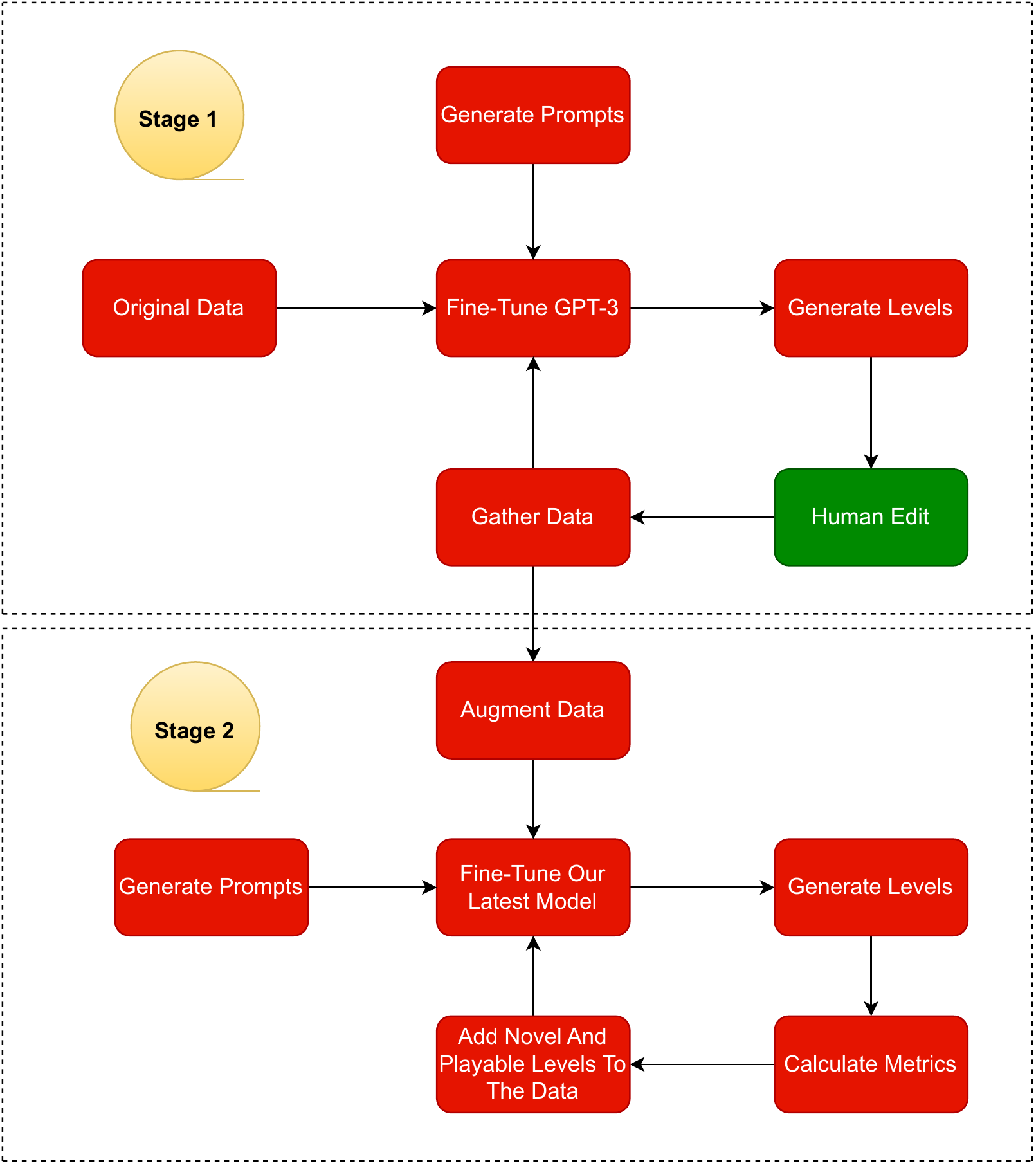}
  \caption{A flowchart of how both stages work. The yellow circle indicates the stage number. Red color indicates automated process and green indicates human-in-the-loop process.}
    \label{fig:flowchart}
\end{figure*}

\subsection{Gameplay}
\textit{Metavoidal}\footnote{https://yellowlabgames.itch.io/metavoidal} is a roguelite brawler game, being developed by \textit{Yellow Lab Games}\footnote{https://yellowlab.games/} that features a metal band trying to hire new drummers. The metal band turns out to be full of eldritch monsters sacrificing drummers to gain more power. You play as a drummer in a church where the auditions are happening. You are trying to escape as you are a bad drummer and hordes of monsters are trying to sacrifice you. You will have drumsticks as your weapon. You can find power-ups like music disks to fight enemies. Your goal is to escape the church.

The layout of the game includes 3 levels. Each level with many rooms connected to hallways. There are some tunnel-like connections. Rooms are essentially the main areas where assets are discovered to progress in the game. Figure \ref{fig:OrigRooms} shows rooms in the game developed by a 2D game artist. There are many tiles to be considered while generating. There are three types of tiles that make patterns: wood, marble and moss. There are two types of walls: marble and moss marble wall. There is one deepwater and one junction point tile. Junction point tile is used for doors. Both of wall tiles and deepwater tiles are considered unwalkable tiles.

\subsection{Constraints For Room Generation}\label{sec:cons}
The following are the constraints: 
\begin{enumerate}
    \item The three pattern tiles that should make the most of the room.
    \item The unwalkable tiles should be at least two tiles apart if it is within the path. The path is defined as all walkable tiles that are connected from one door to another. 
    \item Deepwater tile can be placed in a cluster.
    \item Wall tiles should be placed as a single row of tiles to look like walls.
    \item The walkable tiles should be placed such that one of them should be the base tile and the rest of the two should be supporting pattern tiles.
    \item The junction tile should be one tile apart for vertical doors and two tiles apart for horizontal doors, (5) If there is more than one door then at least two doors should connect to each other.
    \item The length and the width of the room can vary but it should always be divisible by two.
    
\end{enumerate}

\section{Proposed method}

The proposed method takes inspiration from Todd et al.~\cite{todd2023level} and introduces a training technique such that the method can produce playable-novel levels from as scarce data as only 60 rooms. This section will introduce all the methodology in details, starting with the initial dataset, the bootstraping technique~\cite{torrado2020bootstrapping}, augmentation of our dataset, and training of our final model.

\subsection{Dataset}\label{sec:dataset}

We received 60 room levels from the developers as our initial dataset. We map all the rooms to characters as we will give these character-based tiles to our LLM. This is the method followed by Todd et al.\cite{todd2023level} as well. Our LLM is OpenAI's GPT-3. GPT-3 requires prompt and completion as one row of the dataset. As we have many constraints, which is common in any game being developed, we use controllable prompts rather than simply prompting to create a level. Our single prompt looks like this:

\textit{'"The size of the level is \{width x height\}, the base tile is "\{base\_tile\}", and the border tile is "\{border\_tile\}". There are $2$ pattern tiles, "\{pattern\_tiles[0]\}" and "\{pattern\_tiles[1]\}", "F" is the water tile, "J" is the door tile, and the percentage of pattern tiles is \{percent\_pattern\_tiles\}\%.-$>$"'}

Where the \textit{width} and \textit{height} of the level include the border tile. \textit{base\_tile} is the tile with the highest number of counts. \textit{border\_tile} is one of the wall tiles with the most counts on the border. There can be either one or two \textit{pattern\_tiles}, if there is one \textit{pattern\_tiles} then the statement changes to \textit{"There is 1 pattern tile, "\{pattern\_tiles[0]\}"}. \textit{percent\_pattern\_tiles} is calculated by the percentage of only pattern tiles among all tiles. This lets LLM knows how many pattern tiles to use. Lastly, $->$ is used as a special token for GPT-3 to know the prompt has ended. For completions, which are the labels of the prompts, we selected "A", "B" and "C" as walkable tiles, "E" and "\#" as wall tiles, "F" for water tiles and "J" for the junction tile. Each row of tiles ends with \verb|\n| as a newline indicator. After the level ends, we put ". XUT" as the ending token.

\subsection{Augmenting Dataset}

To increase the dataset and get more variation in the dataset, we use the following techniques, motivated by~\cite{todd2023level}, to augment our data:

\begin{enumerate}
    \item We flip the room horizontally and vertically.
    \item We rotate the room $90^{\circ}$ and change the door sizes to cater for the constraint.
    \item We swap the pattern tiles of the original room levels.
    \item We repeat $1 - 2$ for rooms with swapped pattern tiles.
    
\end{enumerate}

\subsection{Our Method}

Our method uses GPT-3 \cite{brown2020language} from OpenAI\footnote{https://openai.com/} as the LLM to generate levels. Reference \cite{todd2023level} shows that \textit{text-davinci-003}, a GPT-3's variant, has the ability to generate Playable-novel levels from scarce data. Our method includes two stages of generation. The first stage is the human-in-the-loop generation where we fine-tune GPT-3 on the data given by developers. In this stage, we get unplayable rooms that do not follow the constraints. We observe all generated rooms and try to fix the ones that are fixable. After fixing the rooms, we add the room to our data if they are novel enough. Torrado et al.~\cite{torrado2020bootstrapping} introduced this method and called it \textit{bootstrapping}, we will continue with the convention. We repeat bootstrapping till we get enough data. We commence the second stage by augmenting our data as explained earlier. Once we have obtained our dataset which is now much larger compared to the initial dataset, we start to generate levels. This stage does not have a human-in-the-loop component. We generate 100 rooms in each round, calculate the metrics, and add the playable-novel rooms back into the dataset. This could essentially be repeated however many times one desires. 

\subsection{Metrics}

Our following two metrics are inspired by Todd et al.~\cite{todd2023level} but adjusted to our needs:

\subsubsection*{Playable-Novel} has two components, playability and novelty. Playability is measured by the constraints mentioned in Section \ref{sec:cons}. Once all the constraints are passed, we check the novelty of the level. We check novelty by first checking if at least the level is novel by a novelty threshold, which is a percentage of the total number of tiles in the created level. This also includes the border tiles. When it is novel by at least the novelty threshold, we swap the pattern tiles and check with the same threshold. The novelty is checked across all the levels in the dataset that we currently have. If it still holds then we consider it Playable-Novel level.

\subsubsection*{Accuracy} is a measure of how close the percentage of generated pattern tiles is to the percentage mentioned in the prompt, and is measured by:

\begin{equation*}
    Accuracy = 1 - \frac{|Prompt\_Percent - Generated\_Percent|}{Prompt\_Percent} 
\end{equation*}

Where $Prompt\_Percent$ is the percentage of pattern tiles, as written in Section \ref{sec:dataset} by $percent\_pattern\_tiles$, and $Generated\_Percent$ is the percentage of the pattern tiles in generated level.

\section{Experiment Setup and Results}

For our first stage of generations, we set the temperature to 0.4. In our experiments, we observe that GPT-3 can generate random text while generating at a higher temperature thus we opted for the specific temperature setting. GPT-3 related settings were set to default. For the novelty score, we set a novelty threshold of $10\%$ of the total number of tiles in the level. As mentioned earlier, We received 60 room levels hand created from the Metavoidal developer team. We generate prompts as discussed earlier. We fine-tune GPT-3 for 5 epochs, generate 100 levels and take the 10 most levels that can be repaired. We repair them and measure their novelty and playability. If novelty and playability are passed, we add them to our data and fine-tune our previously fine-tuned GPT-3 model. We repeat this step till we get 60 more levels. 

\begin{figure}[h!]
    \centering
    \includegraphics[width=0.4\textwidth]{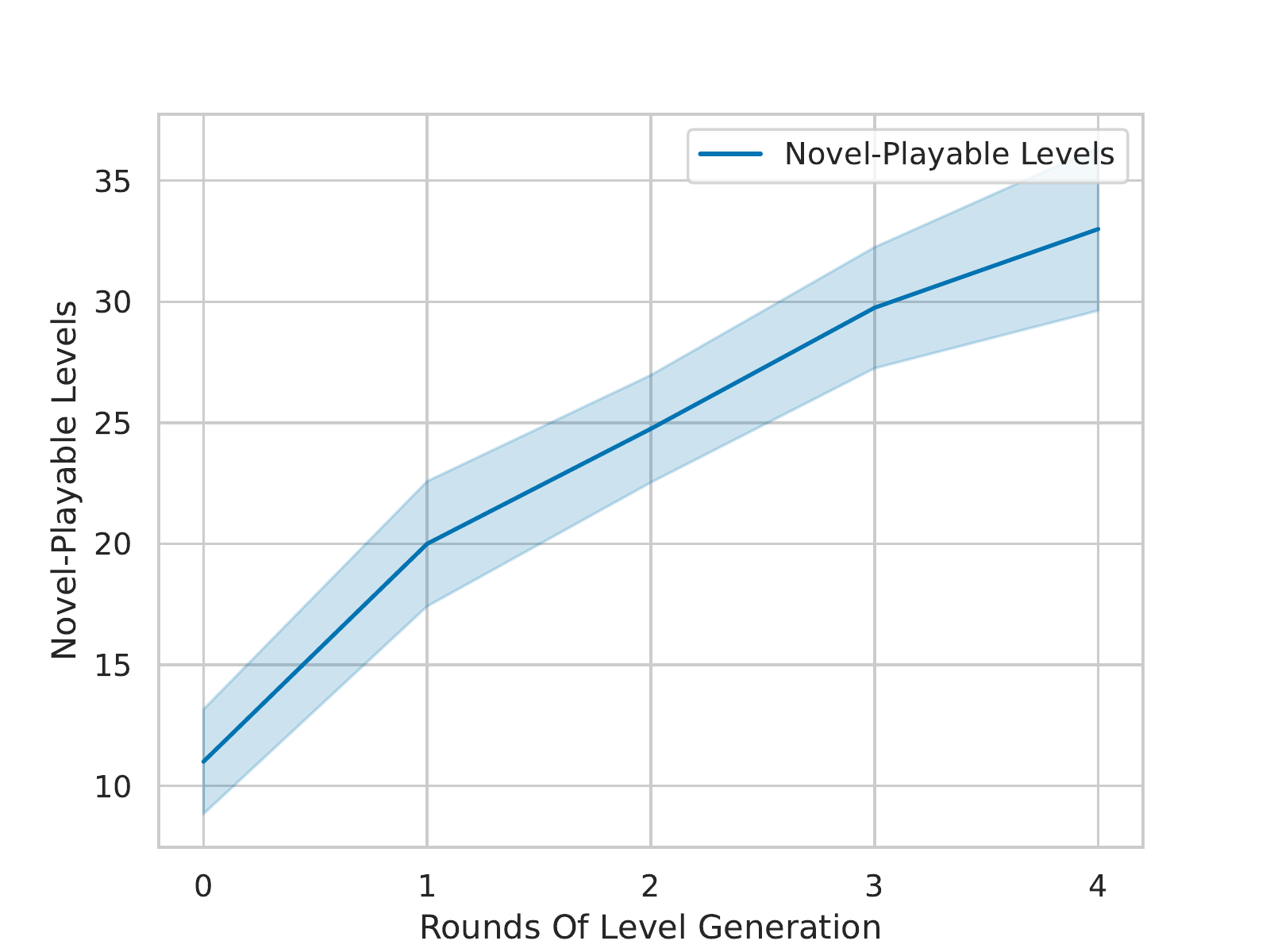}
  \caption{Illustration of how many playable-novel levels are generated per each round of level generation at stage 2. The shadowed region shows variance over 5 seeds while the solid line shows the mean.} 
    \label{fig:novplay}
\end{figure}

With the 120 levels, we move on to the second stage by augmenting our data. We augment our data in the ways we described earlier and obtain 840 levels. We further fine-tune our model for 2 epochs. After fine-tuning, we generate 100 levels to get playable-novel levels. We include them in the dataset to fine-tune our model. We repeat this 5 times to eventually get up to $37\%$ playable-novel levels. Figure \ref{fig:novplay} illustrated what the output of each round of level generation looks like. 

\begin{figure}[h]
    \centering
    \begin{subfigure}{0.13\textwidth}
        \includegraphics[width=\textwidth,keepaspectratio]{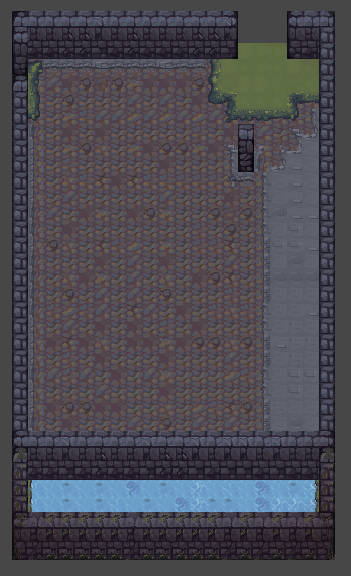}
    \end{subfigure}%
    \begin{subfigure}{0.13\textwidth}
        \includegraphics[width=\textwidth,keepaspectratio]{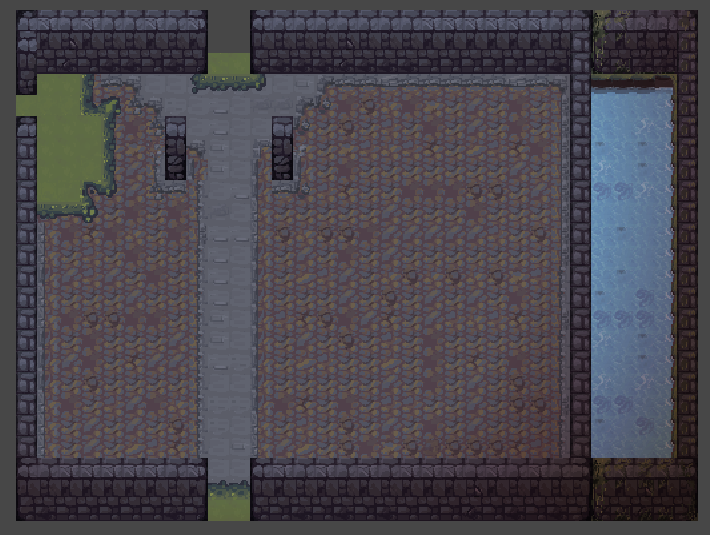}
    \end{subfigure}%
    \begin{subfigure}{0.13\textwidth}
        \includegraphics[width=\textwidth,keepaspectratio]{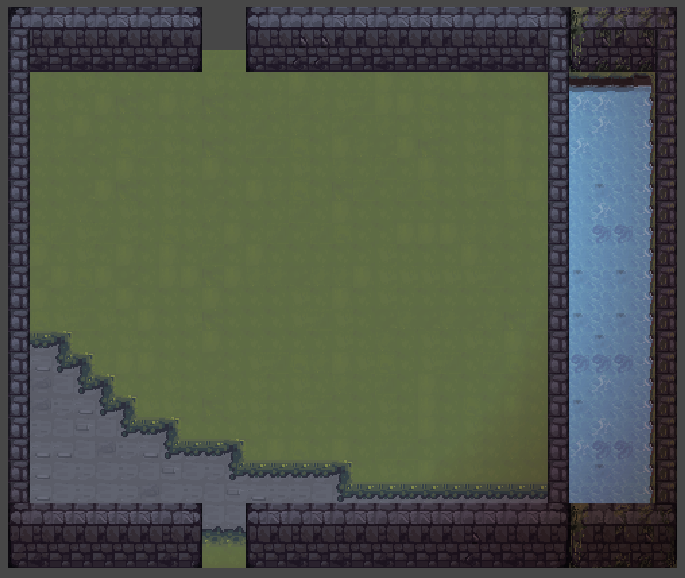}
    \end{subfigure}
    \begin{subfigure}{0.13\textwidth}
        \includegraphics[width=\textwidth,keepaspectratio]{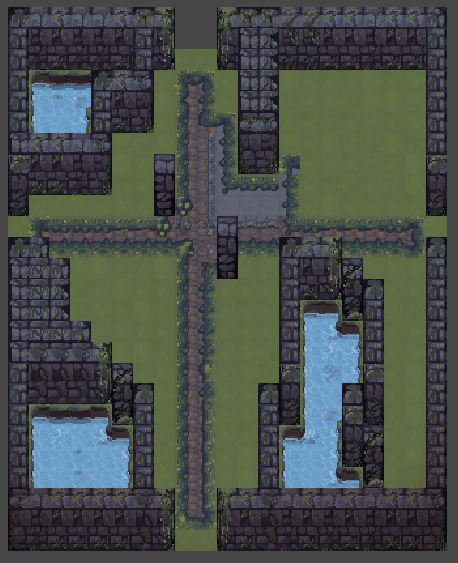}
    \end{subfigure}%
    \begin{subfigure}{0.13\textwidth}
        \includegraphics[width=\textwidth,keepaspectratio]{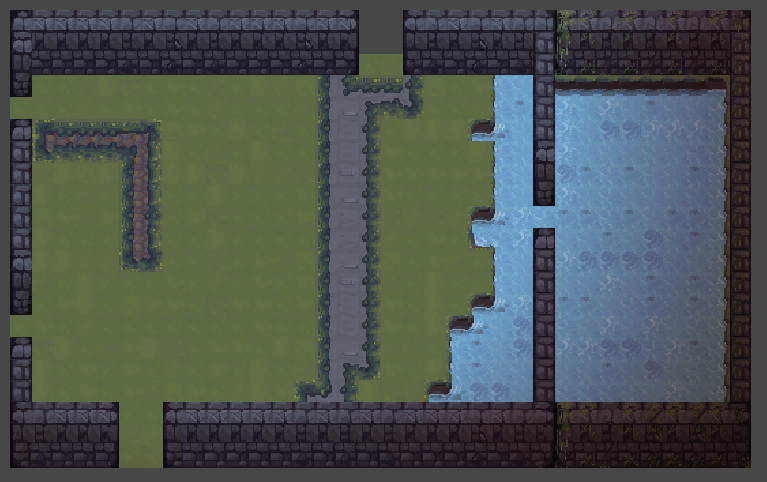}
    \end{subfigure}%
    \begin{subfigure}{0.13\textwidth}
        \includegraphics[width=\textwidth,keepaspectratio]{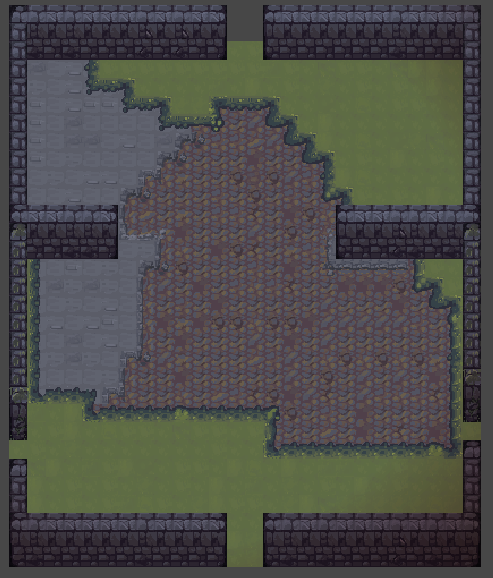}
    \end{subfigure}
    \begin{subfigure}{0.13\textwidth}
        \includegraphics[width=\textwidth,keepaspectratio]{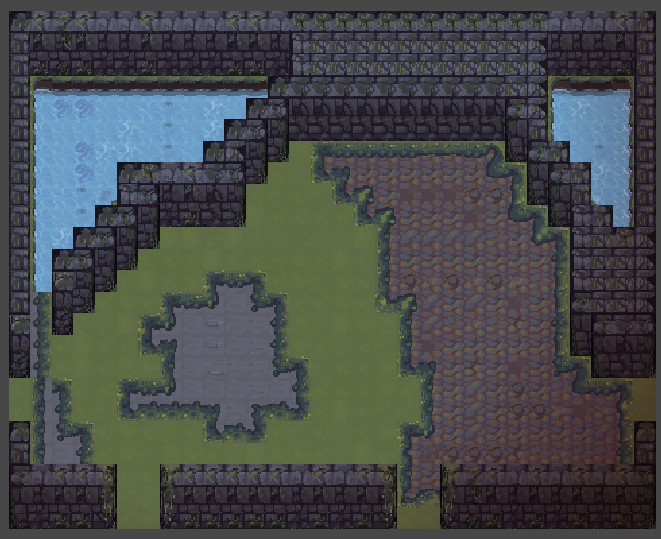}
    \end{subfigure}%
    \begin{subfigure}{0.13\textwidth}
        \includegraphics[width=\textwidth,keepaspectratio]{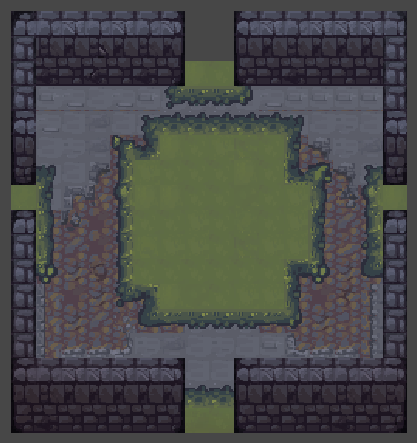}
    \end{subfigure}%
    \begin{subfigure}{0.13\textwidth}
        \includegraphics[width=\textwidth,keepaspectratio]{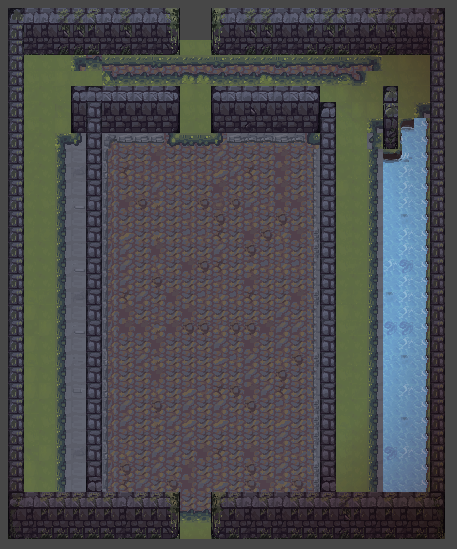}
    \end{subfigure}
    \caption{A few generated rooms, randomly placed, to demonstrate constraints handled by our method.}
\label{fig:genrooms}
\end{figure}

To show the usefulness of the controllability prompt, we illustrate the average accuracy over the generated levels in Figure \ref{fig:genrooms}.
After augmenting the data we implement the rest of the second stage for 5 seeds to solidify our experiments. The total cost for all the experiments was $\$300$.

\begin{figure}[h!]
    \centering
    \includegraphics[width=0.4\textwidth]{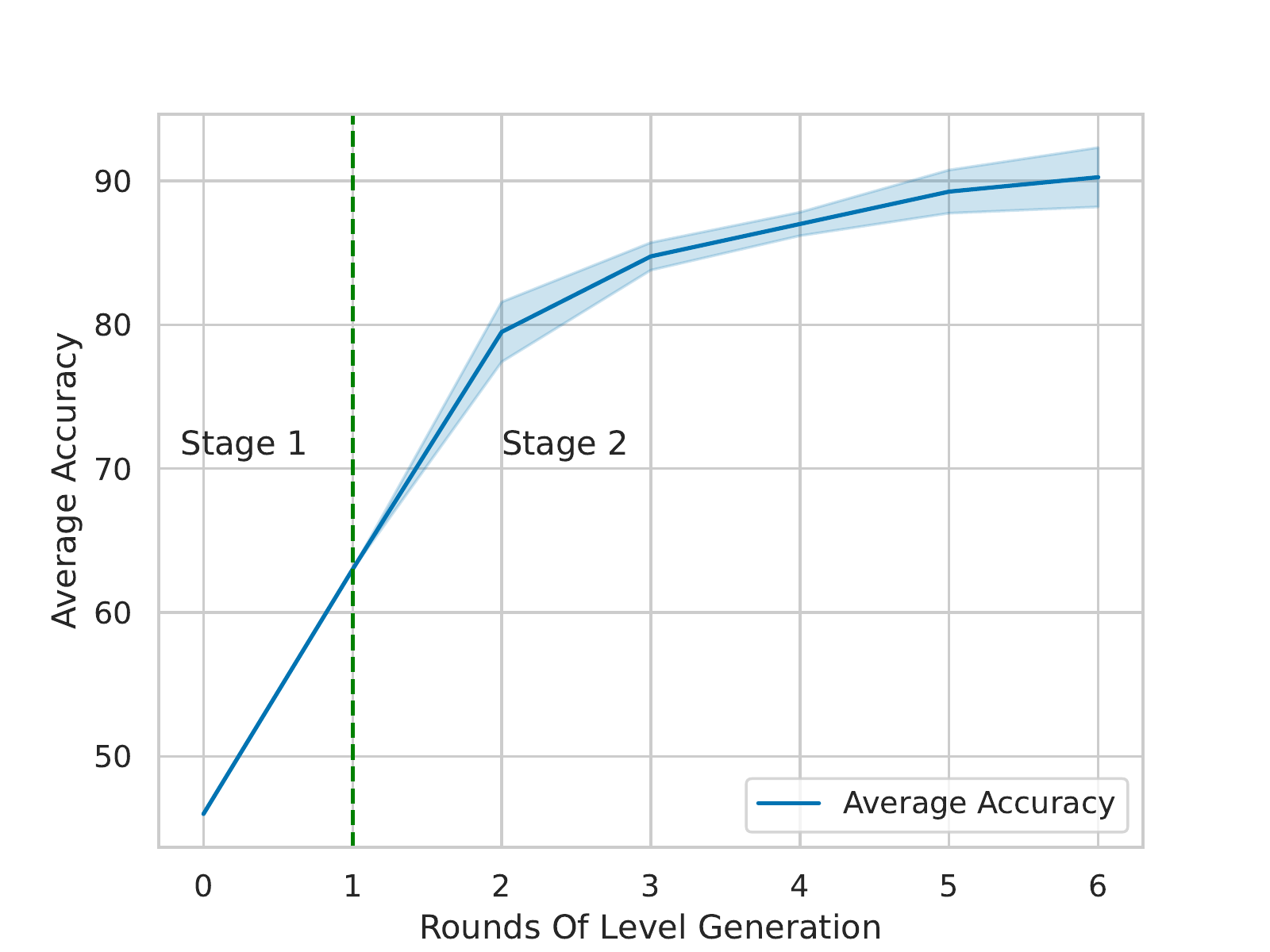}
  \caption{Illustration of how accuracy improves from stage 1 to stage 2 and further to the last round of level generation in stage 2. Stage 1 is performed on 1 seed while stage 2 is performed on 2 seed.} 
    \label{fig:acc}
\end{figure}

\section{Conclusion And Future Directions}

To conclude, we introduce a method on a new aspiring game, named \texttt{Metavoidal}, that can have the maximum characteristics and constraints of the content described in the prompt, while other constraints learnt from scarce data, via GPT-3. Our method is inspired by the work done by the authors of \cite{todd2023level}. We introduce methods of training that lead to the creation of an increasing number of playable-novel levels. We show it on a new game, that is currently under development so that we can introduce the method as a practical tool and an application that can be used to create content in non-trivial games with many constraints. \texttt{Metavoidal} also opens up more space in research as it can be used for PCG, game-playing and game-testing AI research. One of the major future directions for this method is to be working on 3D under-development games. A breakthrough for any kind of PCG via LLM would be to have one generalised model that can generate 2D and 3D levels from a single model.

\bibliographystyle{IEEEtran}
\bibliography{main}

\end{document}